\pdfoutput=1

\documentclass[11pt]{article}

\usepackage{ACL2023}

\usepackage{times}
\usepackage{latexsym}

\usepackage[T1]{fontenc}

\usepackage[utf8]{inputenc}

\usepackage{microtype}

\usepackage{inconsolata}

\usepackage{url}
\usepackage{array}
\usepackage{tabularx}
\usepackage{graphicx}
\usepackage{dblfloatfix}
\usepackage{enumitem}
\usepackage{textcomp}
\usepackage{todonotes}
\usepackage{subcaption}
\usepackage{chngcntr}
\usepackage{linguex}
\usepackage{amsmath}
\usepackage{amssymb}
\usepackage{booktabs}
\usepackage{linguex}
\usepackage{url}
\usepackage{array}
\usepackage{kotex}
\usepackage{multirow}
\usepackage{xcolor}
\newcolumntype{x}[1]{>{\centering\arraybackslash\hspace{0pt}}p{#1}}
\newcolumntype{C}[1]{>{\centering\let\newline\\\arraybackslash\hspace{0pt}}m{#1}}

\newcommand{\textonly}[0]{\textcolor{cyan}{x}}
\newcommand{\textonlytoken}[0]{\textcolor{cyan}{x_t}}
\newcommand{\punct}[0]{\textcolor{olive}{m_t}}
\newcommand{\restored}[0]{\textcolor{magenta}{y}}
\newcommand{\token}[0]{\textcolor{magenta}{y_t}}
\newcommand{\context}[0]{\textcolor{magenta}{y_{<t}}}
\newcommand{\model}[0]{f}
\newcommand{\greentriangle}[0]{\textcolor{green}{$\blacktriangle$}}
\newcommand{\same}[0]{\textcolor{yellow}{$\bullet$}}
\newcommand{\redtriangle}[0]{\textcolor{red}{$\blacktriangledown$}}

\title{Punctuation Restoration Improves Structure Understanding without Supervision}

\author{
 \textbf{Junghyun Min\textsuperscript{1}},
 \textbf{Minho Lee\textsuperscript{2}},
 \textbf{Woochul Lee\textsuperscript},
 \textbf{Yeonsoo Lee\textsuperscript{3}},
\\
 \textsuperscript{1}Linguistics Department, Georgetown University, Washington, DC, USA\\
 \textsuperscript{2}KT Gen AI Lab, Seocho, Seoul, Republic of Korea, \\
 \textsuperscript{3}NC AI, Seongnam, Gyeonggi, Republic of Korea
\\
 \small{
   \textbf{Correspondence:} \href{mailto:jm3743@georgetown.edu}{jm3743@georgetown.edu}}
}

\begin{document}
\maketitle
\begin{abstract}
Unsupervised learning objectives like autoregressive and masked language modeling constitute a significant part in producing pre-trained representations that perform various downstream applications from natural language understanding to conversational tasks. However, despite impressive generative capabilities of recent large language models, their abilities to capture syntactic or semantic structure within text lag behind. We hypothesize that the mismatch between linguistic performance and competence in machines is attributable to insufficient learning of linguistic structure knowledge via currently popular pre-training objectives. Working with English, we show that punctuation restoration as a learning objective improves performance on structure-related tasks like named entity recognition, open information extraction, chunking, and part-of-speech tagging. Punctuation restoration results in \greentriangle$\geq2\%$p improvement in 16 out of 18 experiments, across 6 out of 7 tasks. Our results show that punctuation restoration is an effective learning objective that can improve structure understanding and yield a more robust structure-aware representations of natural language in base-sized models.
\end{abstract}

\newlength{\myindent}
\setlength{\myindent}{0.5cm}

\section{Introduction}
\label{sec:introduction}

The modern natural language processing paradigm centers around transformer-based pre-trained language models (PLMs; \citet{peters-etal-2018-elmo, radford2018gpt, devlin2019bert}). They are optimized on masked language modeling (MLM) and autoregressive language modeling, which provide powerful representations to approach various problems in natural language processing. It is no exaggeration that language models have become effective in tasks like named entity recognition (NER), information extraction, semantic role labeling (SRL) that require understanding of syntactic, semantic, and discourse structure \citep{wang-etal-2021-ace, wang-etal-2022-deepstruct}. However, the following suggests there is still room for improvement in current language models' abilities to understand such structure in natural language to perform downstream tasks reliably and robustly.

\begin{enumerate}
    
\item \textbf{The reversal or factorization curse}. Language models fail to infer "B is A" from "A is B" \citep{berglund2023reversal}, or their representations are highly dependent on the order (factorization) of the input \citep{kitouni2024factorization}.

\item \textbf{The curse of performance instability}. Model checkpoint initialization and training dataset order strongly affects sensitivity to syntactic structure \citep{zhou-etal-2020-curse, mccoy-etal-2020-berts, du-nguyen-2023-measuring}.

\item \textbf{Poor out-of-distribution generalization}. Systems report close-to-human performance on one dataset yet perform poorly on other datasets representing the same task, due to their picking up \textbf{spurious correlations} rather than learning the task \citep{gururangan-etal-2018-annotation, mccoy-etal-2019-right, serrano-etal-2023-stubborn}.

\item \textbf{Insufficient or underutilized structure information}. While PLMs do encode some structure, they are poor few-shot structure predictors \citep{zhao-etal-2023-transformers, bai2023constituency} and perform better when input is reinforced with linguistic structure information \citep{strubell-etal-2018-linguistically, he2020syntactic, sachan2020syntax, wu2021learn, fei-etal-2021-better, xie2023syntax, huang2024flexibly}. This indicates their representations are insufficient or underutilized.
\end{enumerate}

These four phenomena illustrate that current representations as a result of autoregressive \citep{radford2018gpt} or masked \citep{devlin2019bert, liu2019roberta, raffel2019t5} language modeling are insufficient for structure understanding. Efforts to mitigate such shortcomings include data-oriented approaches like syntactic augmentation to improve robustness to spurious correlations \citep{min-etal-2020-syntactic, yaghoobzadeh-etal-2021-increasing} and reversing input to mitigate the reversal curse \citep{golovneva2024reverse}. Architecture oriented efforts include adding explicit graph network layers to encode structure, resulting in improvement in benchmark scores \citep{zhang-etal-2019-syntax, sachan2020syntax, wu2021learn} and generalization abilities \cite{he2020syntactic, sartran-etal-2022-transformergrammar}.

They are human-in-the-loop methods that require human input or annotation, or a system that requires such annotation. Recent work in distilling linguistic structure knowledge from natural language text to representations without supervision include inside-outside dynamic programming for tree induction \citep[DIORA;][]{drozdov-etal-2019-unsupervised-latent}, dependency-constrained self-attention \citep{shen-etal-2021-structformer, momen-etal-2023-increasing}, and augmenting MLM with sentence-level contrastive learning \citep[CLEAR;][]{wu-etal-2020-clear}. With the exception of CLEAR, these methods require additions to the model architecture. \citet{wang-etal-2021-ace} and \citet{wang-etal-2022-deepstruct} propose structure pre-training but use human-annotated data.

In this paper, we investigate whether it is possible for an unsupervised method to mitigate the four shortcomings of the modern language model without implementing additional parser, tree, or graph architecture. In particular, we believe the pre-training stage of current PLMs may be further improved and propose punctuation restoration (PR) as an unsupervised learning objective that improves structure understanding. Punctuation markers, along with capitalization, often serve as boundary markers between different syntactic components of the sentence \citep{briscoe1996syntax, bayraktar1998comma}. Punctuation marks also correspond to prosodic features of the sentence \citep{chafe1988punctuation}, which in turn contain linguistic information \citep{wilson2006relevance, wolf-etal-2023-quantifying}. 
Thus, the model's ability to predict punctuation from plain text may correlate to its ability to encode syntactic boundaries and thus structure. We hypothesize that additional optimization on punctuation restoration yields representations with increased sensitivity to structure, measured by in-distribution test set score, out-of-distribution generalization performance, and stability across initialization in structure-related natural language processing (NLP) tasks. 

Intuitively, punctuation restoration may seem like an easy task. However, predicting punctuation and capitalization given text still remains nontrivial \citep{puaics2022capitalization, DELIMA2024Portuguese, pang2024llama}, and is an area of active work especially for post-processing results from automatic speech recognition and trascription systems \citep{alam2020punctuation, zhu-etal-2024-resolving, you2024light, zhong2025punctuation}. We provide performance on the objective task in Appendix \ref{appendix:obj-results}. Although language modeling already includes predicting punctuation and capitalization, explicit optimization on punctuation restoration would allow models to predict them without explicit local context (e.g. beginning of sentence or quotation, following capitalization).

\section{Objective and experimental setup}
\label{sec:objective_design}

\subsection{Objective design}
\label{subsec:objective_design}

The punctuation restoration objective predicts the original text from its "cleared-formatting" counterpart. In our implementation, we remove the following set of punctuation marks: the comma \textbf{,}, the period \textbf{.}, the exclamation point \textbf{!}, the question mark \textbf{?}, the single-quotation mark \textbf{'}, and the double-quotation mark \textbf{"}, along with capitalization, as shown below. Boldface indicates an addition to or a modification of source text.

\begin{itemize}
\label{fig:faker}
\item Source: lee faker sang-hyeok (hangul: 이상혁) is a league of legends esports player currently mid laner and part owner at t1
\item Target: \textbf{L}ee \textbf{``F}aker\textbf{''} \textbf{S}ang-hyeok (\textbf{H}angul: 이상혁) is a \textbf{L}eague of \textbf{L}egends esports player, currently mid laner and part owner at \textbf{T}1\textbf{.}

\end{itemize}

While it is possible that a different selection yield better results, our selection reflects frequency \citep{sun2019frequency} as well as syntactic significance \citep{bayraktar1998comma, DeBrabanter2023quotation}.

Similarly to popular pre-training objectives like MLM, autoregressive language modeling, and next-sentence prediction, the objective requires no human input. The objective is also architecture-agnostic and can be easily modified as appropriate.

From an internal database of English news articles, accessed between January 2022 and August 2023, we collected a total of 437,031 article excerpts, which are non-overlapping parts separated by a limiting word count of 150. One thousand excerpts each are used as the development and test sets, while the remaining 435,031 excerpts are used for training.

\subsection{Experimental setup}

Our experiments involve two stages. In the first stage, we take the pre-trained weights of the T5-base\footnote{See Appendix \ref{appendix:obj-results} for details on model size selection} model \citep{raffel2019t5}, and perform additional pre-training on the punctuation restoration objective to produce PR-T5. Then, in the second stage, we fine-tune PR-T5 on downstream tasks and datasets.

In the first stage, the model $\model$ is given the "cleared-formatting" token sequence $\textonly$ comprising of tokens $\textonlytoken$ and optimized to predict the original, fully punctuated and capitalized text $\restored$ comprising of tokens $\token$ as described in Section \ref{subsec:objective_design}. However, since there is textual overlap between $\textonly$ and $\restored$, assuming trivial copy error rate, we can write the model $\model$ as a predictor of capitalization and punctuation information $\punct = \token - \textonlytoken$: \[\punct = \model(\textonly, \context) = \begin{cases}
\phi\\
$addPunct$(\textonlytoken, \theta) \\
$addCap$(\textonlytoken, \theta)
\end{cases}\] Thus, the effective loss is as follows: \[
\mathcal{L} \approx -\frac{1}{N} \sum_{t=1}^{N} \log P\left(\punct \mid \textonly, \context\right).
\] In the second stage, we fine-tune PR-T5 and measure the effects of punctuation restoration in downstream tasks. We measure effects across 13 datasets that represent 7 tasks\footnote{See Appendix \ref{appendix:data} for task and dataset details} and across 3 settings: generative, discriminative , and multi-task. In the generative setting, fine-tuned PR-T5 makes entity or tag predictions via autoregressive generation. We conduct 16 experiments in the generative setting, with 13 datasets from 7 tasks. In the multitask setting, fine-tuned PR-T5 is trained to make predictions for two tasks at once, namely NER and Open Information Extraction (OpenIE). We conduct 1 experiment in the multitask setting, with 2 datasets from 2 tasks. Generative and multitask predictions are illustrated in Table \ref{table:multitask-format}. In the discriminative setting, PR-T5's decoder block is replaced with a classification head, as described in Appendix \ref{appendix:discriminative-approach} and Figure \ref{fig:s4e-architecture}. We conduct 1 experiment in the discriminative setting, with 1 dataset from 1 task. We fine-tune the publicly available pre-trained T5 weights on the same downstream tasks and use their performance as comparison baseline for all three settings. We publicly release our \href{https://www.github.com/Aatlantise/punc-rest-improves}{architecture, training, and inference code}.

\section{Results}
\label{sec:results}

We measure the effects of punctuation restoration as an additional pre-training objective on downstream tasks on \texttt{t5-base}, with the four behaviors outlined in Section \ref{sec:introduction} in mind. In this section, we find direct evidence that this method helps mitigate three out of four behaviors we describe in Section \ref{sec:introduction}.

\begin{table*}[h]
    \small
    \centering
    \begin{tabular}{lllccccccc}
    \toprule
         Task & Training set & Evaluation set & \multicolumn{3}{c}{\texttt{t5-base}} & \multicolumn{3}{c}{+ PR} & $\Delta$\\
         \cmidrule(lr){4-6} \cmidrule(lr){7-9} \cmidrule(lr){10-10}
         &&&P&R&F1&P&R&F1&F1 \\
         \midrule
         NER & Econ-mNER & ID & .69 & .65 & .67 & .90 & .89 & .89 & \greentriangle.22\\
         && Econ-sNER & .67 & .76 & .71 & .74 & .81 & .77 & \greentriangle.06\\
         & GENIA & ID & .57 & .73 & .64 & .64 & .76 & .69 & \greentriangle.05\\
         & CoNLL03 & ID & .89 & .90 & .89 & .92 & .92 & .92 & \greentriangle.03\\
         & ontonotes & ID & .87 & .88 & .88 & .91 & .91 & .91 & \greentriangle.03\\ 
         \midrule
         OpenIE & EconIE-PRO & ID & .47 & .43 & .45 & .60 & .63 & .62 & \greentriangle.17\\
         && CaRB & .22 & .16 & .19 & .62 & .42 & .50 & \greentriangle.31 \\
         & OIE2016 & ID & .16 & .19 & .18 & .19 & .19 & .19 & \same.01\\ 
         && CaRB & .10 & .15& .12 & .26 & .27 & .27 & \greentriangle.15 \\
         \midrule
         Chunking & CoNLL00 & ID & .94 & .94 & .94 & .96 & .96 & .96 & \greentriangle.02\\
         && CoNLL03 & .41 & .41 & .41 & .41 & .42 & .42 & \same.01\\ \midrule
         SRL & CoNLL12 & ID & .75 & .79 & .77 & .84 & .86 & .85 & \greentriangle.08\\
         \midrule
         SBD & PTB & ID & .97 & .72 & .81 & .98 & .98 & .98 & \greentriangle.17\\
         \midrule
         POS & CoNLL00 & ID & .96 & .96 & .96 & .98 & .98 & .98 & \greentriangle.02 \\
         && CoNLL03 & .74 & .87 & .79 & .84 & .88 & .86 & \greentriangle.07 \\ \midrule
         RE & TACRED & ID & & & .67 & & & .83 & \greentriangle.16 \\
         \bottomrule      
    \end{tabular}
    \caption{Our main results where we compare \texttt{t5-base} model to \texttt{PR-t5-base} (+PR). ID denotes in-distribution evaluation on a dataset from the same source as the training set. See Appendix \ref{appendix:data} for dataset details.}
    \label{table:generative-results}
\end{table*}

\begin{table}[t]
    \small
    \centering
    \resizebox{\columnwidth}{!}{
    \begin{tabular}{lccccccc} \toprule
        & \multicolumn{3}{c}{\texttt{t5-base} (joint)} & \multicolumn{3}{c}{+ PR} & $\Delta$\\
        \cmidrule(lr){2-4} \cmidrule(lr){5-7} \cmidrule(lr){8-8}
         & P & R & F1 & P & R & F1 & F1 \\
        \midrule
        NER & .86 & .84 & .85 & .87 & .86 & .87 & \greentriangle.02\\
        OIE    & .57 & .60 & .58 & .60 & .62 & .61 & \greentriangle.03\\
         \bottomrule
    \end{tabular}}
    \caption{Multitask (Econ-mNER, EconIE-PRO) performance.}
    \label{table:multitask-results}
\end{table}

\begin{table}[t]
    \centering
    \resizebox{\columnwidth}{!}{
    \begin{tabular}{lccccccc} \toprule
        & \multicolumn{3}{c}{\texttt{t5-base} (EO)} & \multicolumn{3}{c}{+ PR} & $\Delta$\\
        \cmidrule(lr){2-4} \cmidrule(lr){5-7} \cmidrule(lr){8-8}
        & P & R & F1 & P & R & F1 &F1\\
        \midrule
        min & .67 & .91 & .78 & .74 & .90 & .82 & \greentriangle.04\\
        max & .88 & . 94 & .91 & .90 & .94 & .91 & \same.00\\
        avg & .78 & .93 & .85 & .83 & .92 & .88 & \greentriangle.03\\
        sdev & .061 & .009 & .035 & .048 & .010 & .027 & \redtriangle.008 \\
         \bottomrule
    \end{tabular}}
    \caption{Discriminative Econ-mNER performance.}
    \label{table:discriminative-results}
\end{table}

We report our results in Tables \ref{table:generative-results}, \ref{table:multitask-results}, \ref{table:discriminative-results}. Each reported value of precision, recall, and F1 represents an average over the same 5 seed initializations, with the exception of discriminative NER, where we analyze 15 seed initializations.

\subsection{Structure information encoding and use}
In all 18 experiments across dataset, task, and setting, PR-T5 reports improved performance over T5 baselines. Among them, 16 experiments report improvements \greentriangle $\geq.02$, and 10 experiments \greentriangle $\geq.05$ (Tables \ref{table:generative-results}, \ref{table:multitask-results}, \ref{table:discriminative-results}). This is evidence that punctuation restoration makes available a nontrivial amount of structure information that previously may have been unavailable or underutilized, mitigating behavior 4 from Section \ref{sec:introduction}.

\subsection{Performance stability and out-of-distribution generalization}

An out-of-distribution evaluation measures performance on a dataset that represents the same task but comes from a different source than the training dataset (e.g. evaluating on CaRB \citep{bhardwaj-etal-2019-carb} after fine-tuning on OIE2016 \citep{stanovsky-dagan-2016-oie2016}). It is an effective measure of robustness of a representation, as fine-tuned models often learn the dataset, rather than learning the task \citep{gururangan-etal-2018-annotation, mccoy-etal-2019-right, serrano-etal-2023-stubborn}. We compare out-of-distribution generalization ability of PR-T5 to that of T5 in 5 experiments across NER, OpenIE, Chunking, and POS tagging, where we observe \greentriangle$\geq.05$ increase in 4 of them (Table \ref{table:generative-results}). This is evidence that punctuation restoration improves out-of-distribution generalization, mitigating behavior 3 in Section \ref{sec:introduction}.

In addition, we observe that punctuation restoration reduces performance instability. Compared to T5, PR-T5's distribution of NER performance across initialization seeds is narrower. Minimum-maximum range (\redtriangle.04) and standard deviation (\redtriangle23\%) both decrease with additional pre-training in punctuation restoration, as reported in Table \ref{table:discriminative-results}. The results support our hypothesis that punctuation restoration increases stability across initialization seed and training dataset order, mitigating behavior 2 discussed in Section \ref{sec:introduction}.

\section{Discussion}
\label{sec:discussion}
Results from Section \ref{sec:results} support our hypothesis that complementing MLM with a more structure-related objective improves structure understanding. In particular, we use a punctuation restoration objective, described in Section \ref{sec:objective_design} and evaluate with various structure-related tasks. While it is difficult to investigate the exact mechanism of how additional training on punctuation restoration improves learned representations, we attempt to provide an explanation.

In Section \ref{sec:introduction}, we analyze that current methods for representation learning during the pre-training stage lack sufficient signal, and hypothesize additional training with a structure-sensitive objective should improve structure understanding. Much like how prosody helps disambiguate syntax in human speech processing \citep{price1991use, kahn2005effective}, punctuation can be a useful guide in syntax disambiguation, and eventually toward forming a robust representation of text. Punctuation marks represent prosodic design \citep{chafe1988punctuation} which carries linguistic information \citep{wilson2006relevance, wolf-etal-2023-quantifying}. Punctuation marks often also indicate syntactic or semantic boundaries \citep{briscoe1996syntax, bayraktar1998comma}. Optimizing a computational system to predict punctuation allows it to predict syntactic and semantic boundaries, even in the absence of punctuation in the original text. Sufficient training in restoring punctuation can imitate effects of explicitly providing a parse, facilitating natural language understanding via a stronger understanding of sentence structure.

Performance improvement from punctuation restoration is not limited to a specific dataset, task, and setting\footnote{And decoding method, discussed in Appendix \ref{appendix:data}}. and represents an overall increase in representation robustness, as we observe out-of-distribution performance jump in NER, OpenIE, and chunking. Because of the wide range of experiments in which improvement is observed, we interpret this to be a general improvement of structure understanding rather than fortunate task-specific artifacts from the additional training.

Our methods yield a more reliable and robust representation that can be easily implemented and do not interfere with architectural additions. Punctuation restoration can be applied to reinforce structure understanding and improve robustness of learned representations regardless of model choice, or task-specific engineering policy. The effective objective requires no supervision, and one can construct a training corpus with little computational or manual resources.

\section*{Limitations}

The idea of structure understanding reinforcement via punctuation restoration is still young--many decisions relevant to the learning objective in this paper, including selection of punctuation marks and source of learning corpus warrant additional investigation in future work. Our set of training hyper-parameters also will benefit from additional attention.

Among the 4 behaviors discussed in Section \ref{sec:introduction}, we find direct evidence that punctuation restoration mitigates only three of them. While we predict that unsupervised structure learning via objectives like punctuation restoration can help mitigate the reversal (factorization) curse, this will need explicit verification.

While our experiments show promise in base-sized natural language understanding models for English, its effects in larger models, implications to generative or conversational systems, and generalization to other languages and thus language-agnostic nature also need to be verified.

It is also likely that punctuation restoration is not the only unsupervised learning objective that can be used to improve the representation learning stage of training NLP systems. Other forms of unsupervised structure learning, possibly simpler and more effective methods than punctuation restoration, as well as optimizations on objective combination (e.g. with word prediction methods) should be studied in future work.

Finally, although we believe that additional optimization on punctuation restoration improves the models' encoded linguistic structure, leading to performance jumps between T5 and PR-T5, we do not control for the additional compute or the exposure to novel data in this paper. Future work explicitly controlling for such variables will provide more robust arguments for punctuation restoration as a representation learning objective.

\section*{Responsible research statement}

We use OpenAI's GPT-3.5 Turbo \citep{brown-2020-gpt3} as a punctuation restoration performance baseline, and as a debugging assistant during the project's technical implementation.

The Econ-mNER dataset was annotated by paid, full-time employees who are trained linguists knowledgeable about their work and the dataset's downstream use. They are compensated similarly to the region's 2021 median income level. Their work has been reviewed by an internal board to not contain any personally identifiable information. Other internal datasets did not require manual annotation.

\section*{Acknowledgements}

This work was performed during the authors' time at NC AI. We thank Yerang Kim, Tatsuya Aoyama, and Ethan Wilcox for their helpful comments. Any errors remain our own.

\bibliography{custom}
\bibliographystyle{acl_natbib}

\appendix
\section{Additional details on experimental setup}
\label{appendix:exp}

We train the model on the punctuation restoration objective for 40 epochs, before fine-tuning with supervised datasets for downstream tasks. The experiments are run on a single V100 GPU with 32GB VRAM, with half precision and gradient accumulation enabled at 16. Our choice of hyper-parameters are as follows: batch size 32, maximum sequence length 256, learning rate 3e-4, maximum grad norm 0.5, and Adam epsilon 1e-8. Number of fine-tuning epochs was 10, with the exception of SRL, which is fine-tuned for 1 epoch only. The additional pre-training lasts about 2 weeks, while the length of each epoch of training varies across datasets between 10 minutes and around 2 hours.

\begin{figure*}[h]
    \centering
    \includegraphics[width=5.5in]{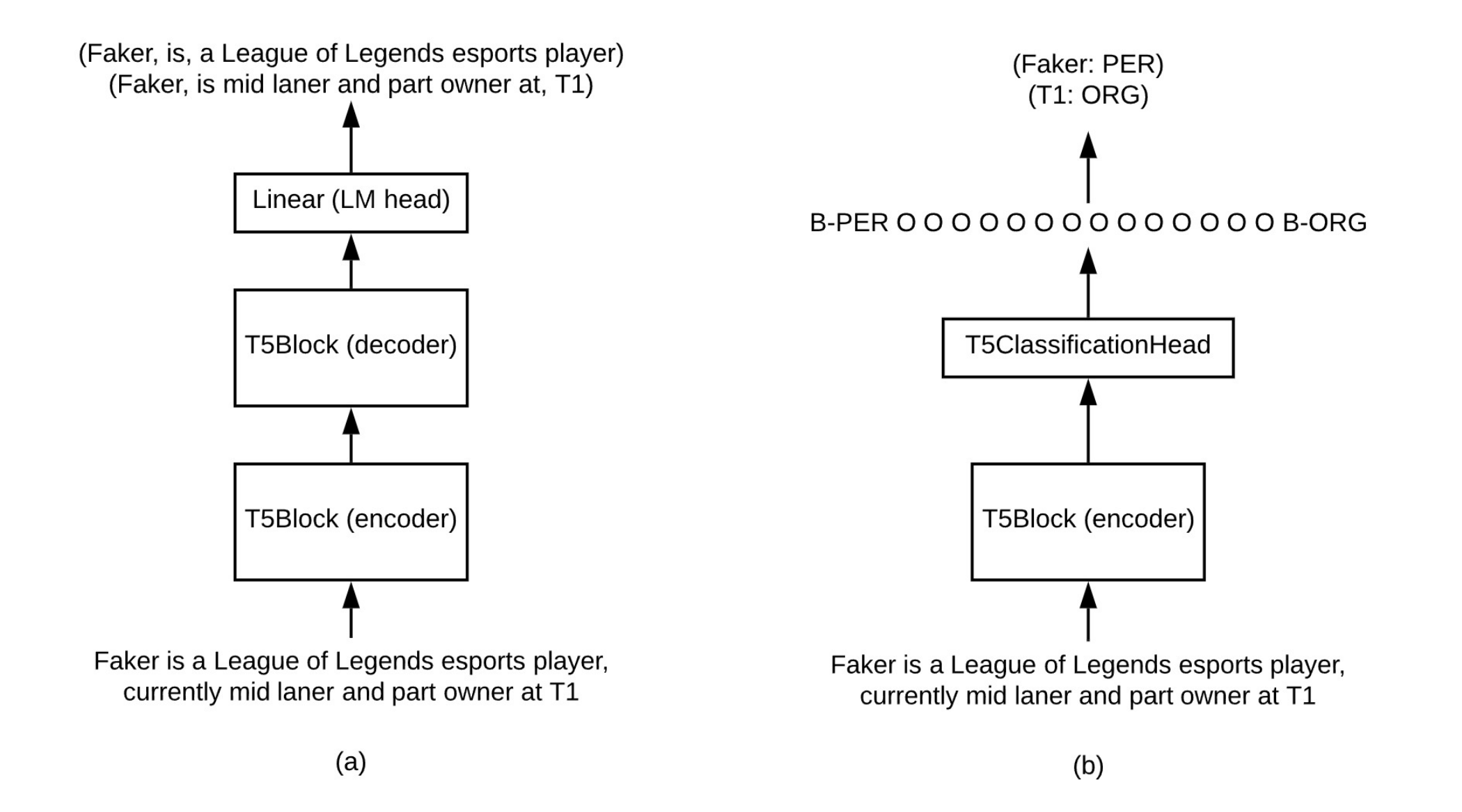}
    \caption{(a) The \texttt{t5} architecture for a generative, text-to-text approach to NLP tasks. Here, we illustrate open information extraction. (b) A modification to the \texttt{t5} architecture to allow a discriminative approach to NLP tasks. Here, we illustrate named entity recognition.}
    \label{fig:s4e-architecture}
\end{figure*}

\subsection{Discriminative approach}
\label{appendix:discriminative-approach}

While there exist sophisticated attempts to incorporate the decoder layers in producing a discriminative model from a pre-trained encoder-decoder architecture \citep{liu2022enct5}, we use a simple architecture where we forgo the decoder block and place a \texttt{T5ClassificationHead} on top of the encoder block of the T5 model. That is, we take the hidden state output from model's encoder and use it as input to the classification head. An illustration of the model architecture is shown in Figure \ref{fig:s4e-architecture}. After additional pre-training on punctuation restoration objective, the decoder block of the \texttt{t5-base} model is removed and a newly initialized classification head is placed on top of the encoder block. The architecture is comparable to those of BERT-like encoder-only models. Even by retaining weights from the encoder blocks only, we observe that additional unsupervised structure learning via punctuation restoration results in downstream task performance improvement.

\subsection{Joint multitask generative approach}
The joint multitask approach, where we focus on open information extraction using the EconIE-PRO dataset and NER using the Econ-mNER dataset, is similar to the generative approach. The input sequence is identical to the experiments from Section \ref{sec:results}, but the output sequence is a concatenation of output sequences from the two datasets, as illustrated in Table \ref{table:multitask-format}.

\section{Additional details on dataset}
\label{appendix:data}

We use a suite of structure-related NLP tasks to measure model structure understanding. Relevant tasks include named entity recognition (NER), sentence boundary detection (SBD), open information extraction (OpenIE), chunking, semantic role labeling (SRL), part-of-speech tagging, and relation classification. Our selection mostly follows that from \citet{wang-etal-2021-ace} and \citet{lee-etal-2024-structured}. We use both public and internal datasets, and check for in- and out-of-distribution generalization. A full list of datasets for each task is shown in Table \ref{table:datasets}. In the main body of the paper, we discuss effects of punctuation restoration across task, dataset, and setting. Here, we discuss another variable across which punctuation restoration is effective: decoding method.

\subsection{Entity generation tasks}
NER, OpenIE, SRL, and relation classification are entity generation tasks, where fine-tuned models autoregressively generate entity objects. For example, \texttt{(Faker: PER), (Faker, is, a League of Legends esports player), (Faker, employeeAt, T1)} are NER, OpenIE, and relation classification examples, respectively. The order in which entities are generated does not affect evaluation in the case of entity generation tasks.

\begin{table}[h]
    \centering
    \resizebox{\columnwidth}{!}{
    \begin{tabular}{ll} \toprule
    \textbf{Source} & Faker is a League of Legends esports player,\\ & currently mid laner and part owner at T1. \\
    \midrule
    \textbf{OpenIE} & (Faker, is, a League of Legends esports player) \\ & (Faker, is mid laner and part owner at, T1) \\
    \textbf{NER} & (Faker: PER) (T1: ORG) \\
    \textbf{Multitask} & (Faker: PER) \\ & (Faker, is, a League of Legends esports player) \\
    & (Faker, is mid laner and part owner at, T1) \\ & (T1: ORG) \\
    \bottomrule
    \end{tabular}}
    \caption{Example output from generative NER, OpenIE, and multitask models.}
    \label{table:multitask-format}
\end{table}

\subsection{Tag sequence generation tasks}
Chunking and POS tagging are tag sequence generation tasks, where fine-tuned models auto-regressively generate tag sequences. \texttt{"NP VP ADVP PP NP NP NP"} and \texttt{"NP VBZ DT NP IN NP"} are example sequences of chunking and POS tagging, respectively.

\subsection{Sequence generation tasks}
Punctuation restoration and sentence boundary detection are sequence generation tasks. Fine-tuned models auto-regressively generate natural text sequences, with predefined tags to perform the task. For example, a sentence boundary detection model would generate a \texttt[<s>] token between sentences, given a passage.

\begin{table*}[th]
    \centering
    \resizebox{\textwidth}{!}{
    \begin{tabular}{clll} \toprule
         \multicolumn{1}{c}{Task} & \multicolumn{1}{c}{Dataset} & \multicolumn{1}{c}{Source} & Task type \\ \midrule
         \multicolumn{2}{l}{\textbf{Internal datasets}} &&  \\ \midrule
         PR & finPR & Rule-based tagging on finance news & Seq. gen.\\ 
         NER & Econ-mNER & Manual tagging on finance news and corporate filings & Ent. gen., Tok. cls.\\
         & Econ-sNER & Semi-supervised tagging on finance news & Ent. gen. \\ 
         OpenIE & EconIE-PRO & Rule-based tagging on finance news, predicate range optimized & Ent. gen.\\
         \midrule
         \multicolumn{2}{l}{\textbf{Public datasets}} \\ \midrule
         NER & GENIA & \citet{kim2003genia} & Ent. gen.\\
         & CoNLL 2003 & \citet{tjong-kim-sang-de-meulder-2003-conll2003} & Ent. gen.\\
         & ontonotes & \citet{weischedel-2013-ontonotes} & Ent. gen.\\
         SBD & PTB & \citet{marcus-1993-ptb} & Seq. gen.\\
         OpenIE & OIE2016 & \citet{stanovsky-dagan-2016-oie2016} & Ent. gen.\\
         & CaRB & \citet{bhardwaj-etal-2019-carb} & Ent. gen.\\
         Chunk, POS & CoNLL 2000 & \citet{tjong-kim-sang-buchholz-2000-conll2000} & Tag gen.\\
         & CoNLL 2003 & \citet{tjong-kim-sang-de-meulder-2003-conll2003} & Tag gen.\\
         SRL & CoNLL 2012 & \citet{pradhan-etal-2012-conll2012} & Ent. gen.\\
         ORE & TACRED & \citet{zhang2017tacred} & Ent. gen.\\
         \bottomrule
    \end{tabular}}
    \caption{We use a total of 14 datasets across 8 tasks, including punctuation restoration. Four are internal datasets, while the rest are publicly available.}
    \label{table:datasets}
\end{table*}

\subsection{Token classification tasks}
NER in the discriminative setting is a token classification task. Given a sentence of length $n$, the fine-tuned model outputs an array of length $n$, each element of which represents whether its corresponding token is part of a named entity. For example, one from a tag set such as \texttt{[O, B-PER, I-PER, B-LOC, I-LOC, B-ORG, I-ORG]}, as illustrated in Figure \ref{fig:s4e-architecture}.

\begin{table}
    \centering
    \begin{tabular}{lccc} \toprule
        Model architecture & P & R & F1 \\
        \midrule
        ChatGPT 0-shot* & .75 & .71 & .73 \\
        \texttt{t5-small} & .91 & .86 & .88 \\
        \texttt{t5-base} & .93 & .92 & .93 \\
        \texttt{t5-large} & .94 & .93 & .93 \\
         \bottomrule
    \end{tabular}
    \caption{Punctuation restoration performance after 50 epochs (small), 40 epochs (base), and 20 epochs (large) of training respectively. *Measured on a small subset of the punctuation restoration evaluation dataset.}
    \label{table:punctuation-restoration-performance}
\end{table}

\section{Additional details on results}
\label{appendix:res}

In our results, improvements from punctuation restoration persist across decoding methods--entity generation in NER, OpenIE, SRL, and relation classification; tag sequence generation in chunking and POS tagging; sequence generation in sentence boundary detection; and token classification in discriminative NER. 

\subsection{Objective results}
\label{appendix:obj-results}

Punctuation restoration, along with capitalization restoration is no trivial task, especially when the model needs to predict restoration location without local context \citep{gravano-2009-restoring, alam2020punctuation, puaics2022capitalization, DELIMA2024Portuguese, zhong2025punctuation}. Should our hypothesis hold, it is likely that syntactic signals from punctuation restoration transfer more effectively in models with stronger punctuation restoration performances. We experiment with three sizes of the T5 architecture \citep{raffel2019t5}. We consider \texttt{t5-small}, \texttt{t5-base}, and \texttt{t5-large}. Table \ref{table:punctuation-restoration-performance} includes their punctuation restoration performance, in addition to ChatGPT's \citep{brown-2020-gpt3} zero-shot performance as a reference point, which shows that the objective is nontrivial.

Across the T5 models, there is some correlation between size and punctuation restoration performance. Because the performance gap between \texttt{t5-base} and \texttt{t5-large} models is small (\same.00), while gap between \texttt{t5-small} and \texttt{t5-base} more significant (\greentriangle.05), we use the \texttt{t5-base} model for our experiments.

We also note that our selection of the T5 model is due to its ability to perform both generative and discriminative tasks after single pre-training.

\subsection{Joint multitask generative setting}
\label{sec:multitask-results}

Similarly to the generative approach, we observe that additional unsupervised structure learning via punctuation restoration results in downstream task performance improvement (\greentriangle.02 NER and \greentriangle.03 OpenIE). While PR-T5 multi-task performance slightly degrades compared to its single-task generative setting (\redtriangle .02 NER and \same.01 OpenIE), multitask-T5 outperforms single task-T5 on EconIE-PRO, an open information extraction dataset (\greentriangle.13).

\subsection{Discriminative setting}
\label{sec:discriminative-results}

Given the results from the single-task generative approach, the transfer from punctuation restoration to multi-task generative approach may be no big surprise, as there is no drastic difference between the generative nature of the two approaches. However, we report that our improved representations from punctuation restoration non-trivially transfers to the discriminative approach as well, where the decoder block is removed from the model, as illustrated in Figure \ref{fig:s4e-architecture}. Although the maximum performance for T5 and PR-T5 are similar at .91 (\same.00), there is a significant difference in the minimum, at .78 and .82, respectively (\greentriangle.04). Punctuation restoration results in not only higher performance, but also more consistent and stable sets across different initializations.

\end{document}